\documentclass{article} 
\usepackage[preprint]{preprint}

\usepackage{microtype}
\usepackage{hyperref}
\usepackage{url}
\usepackage{booktabs}

\usepackage{graphicx}
\usepackage{subcaption}
\usepackage{multirow}
\usepackage{amsmath}
\usepackage{amssymb}
\usepackage{mathtools}
\usepackage[capitalize,noabbrev]{cleveref}
\usepackage[ruled,linesnumbered,vlined]{algorithm2e}

\usepackage[most]{tcolorbox}
\newtcolorbox{highlight}[1][]{
    enhanced,
    colback=yellow!10,
    colframe=gray!30,
    boxrule=0.5pt,
    arc=2pt,
    leftrule=3pt,
    rightrule=3pt,
    toprule=1pt,
    bottomrule=1pt,
    breakable,
    #1
}
\makeatletter
\def\blfootnote{\xdef\@thefnmark{}\@footnotetext}
\makeatother

\usepackage{bm}
\usepackage{bbm}
\usepackage[mathscr]{eucal}

\newcounter{subassumption}[asu]

\def \endprf{\hfill {\vrule height6pt width6pt depth0pt}\medskip}

\def\1{\bm{1}}

\def\vh{{\bm{h}}}

\def\vx{{\bm{x}}}

\DeclareMathAlphabet{\mathsfit}{\encodingdefault}{\sfdefault}{m}{sl}
\SetMathAlphabet{\mathsfit}{bold}{\encodingdefault}{\sfdefault}{bx}{n}

\SetKwInput{KwInput}{Input}
\SetKwInput{KwOutput}{Output}
\SetKwFunction{FAdaptiveSelect}{AdaptiveSelect}
\SetKwFunction{FWindowAlign}{WindowAlignedPad}
\SetKwFunction{FBuildSequence}{BuildSequence}
\SetKwFunction{FProcessExample}{ProcessExample}
\SetKwProg{Fn}{Function}{:}{}
\newcommand{\Append}[2]{\tilde{\mathbf{c}} \mathrel{{+}{=}} #1,\quad \mathbf{m} \mathrel{{+}{=}} #2}

\usepackage{lineno}

\definecolor{darkblue}{rgb}{0, 0, 0.5}
\hypersetup{colorlinks=true, citecolor=darkblue, linkcolor=darkblue, urlcolor=darkblue}

\title{SuperThoughts: Reasoning Tokens in Superposition}

\author{
Zheyang Xiong$^{w,m}$, Shivam Garg$^{*m}$, Max Yu$^{*i}$, Vaishnavi Shrivastava$^{m}$, Haoyu Zhao$^{p,m}$\\
\textbf{Anastasios Kyrillidis}$^{r}$, \textbf{Dimitris Papailiopoulos}$^{w,m}$ \\
  $^w$University of Wisconsin-Madison, $^m$Microsoft Research, $^i$Independent\\$^p$Princeton University, $^r$Rice University
}

\begin{document}

\ifcolmsubmission
\linenumbers
\fi

\maketitle

\begin{abstract}
Long Chain-of-Thought (CoT) reasoning improves LLM problem-solving but is computationally expensive due to sequential token generation. While recent works explore reasoning in continuous latent spaces to bypass discrete token generation, they often struggle with training stability and fail to scale to complex, long-horizon tasks due to lack of supervision signal. We propose SuperThoughts, which compresses pairs of consecutive CoT tokens into single latent representations and decodes two tokens per step via a lightweight Multi-Token Prediction (MTP) module. This preserves discrete token supervision at training time while doubling throughput at inference time. We finetune Qwen2.5-Math-1.5B-Instruct, Qwen2.5-Math-7B-Instruct, Qwen2.5-Math-14B-Instruct, and evaluate on MATH500, AMC, OlympiadBench, and GPQA-Diamond. With a confidence-based adaptive mechanism that falls back to standard decoding when uncertain, SuperThoughts achieves $\sim$20--30\% CoT length reduction while maintaining accuracy with minimal degradation (1-2 points accuracy drop on most tasks).
\begin{figure*}[h]
    \centering
    \includegraphics[width=\linewidth]{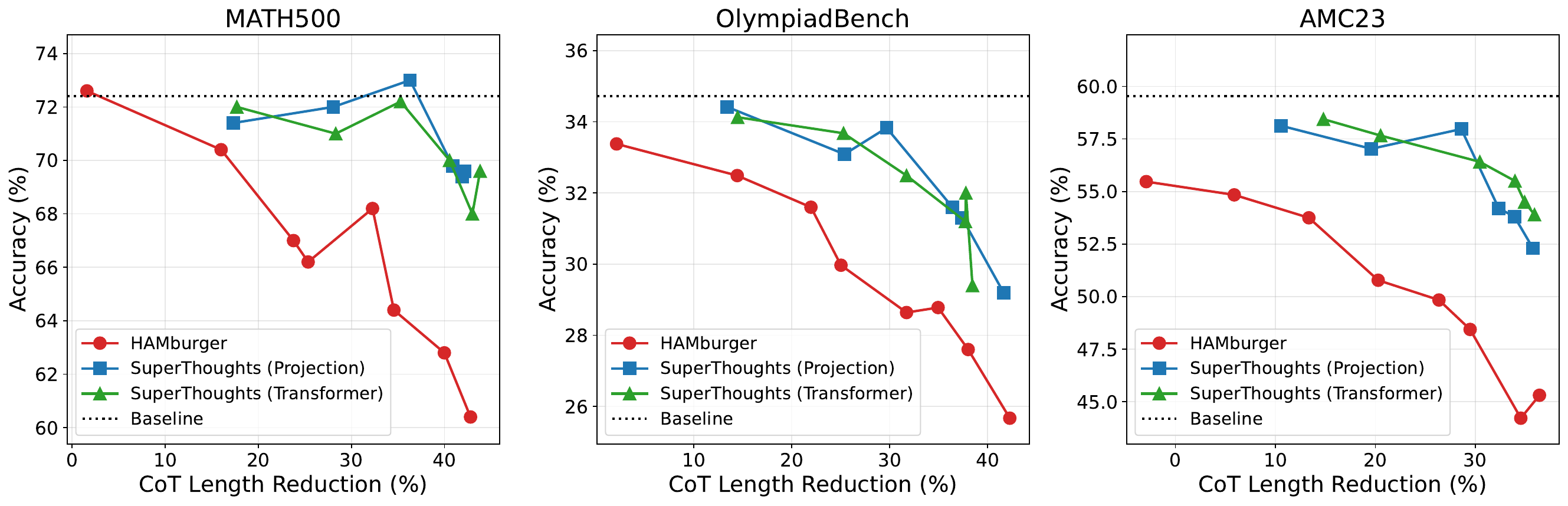}
    \caption{Comparison between SuperThoughts and HAMburger \citep{liu2025hamburgeracceleratingllminference} on trained Qwen2.5-1.5B-Math-Instruct.}
    \label{fig:hamburger_comp}
\end{figure*}
\end{abstract}

\blfootnote{$^*$Equal contribution. Email: \texttt{<zheyang@cs.wisc.edu>}. Correspondence: \texttt{<dimitris@papail.io>}.}

\section{Introduction}

Large language models (LLMs) solve complex problems by generating explicit Chain-of-Thought (CoT) sequences before arriving at a final answer~\citep{wei2022chain}. We can view each CoT token as a unit of compute (one forward pass), and longer chains mean more computation spent before reaching the answer. Recent successes such as OpenAI o1~\citep{jaech2024openai} and DeepSeek-R1~\citep{guo2025deepseek} demonstrate that this additional test-time compute substantially improves performance~\citep{snell2024scaling}.

This raises a question: \emph{why must the model reason in discrete token space?} The vocabulary of a language model is a finite, human-interpretable set of symbols, yet the model's internal representations live in a continuous, high-dimensional vector space. If reasoning could occur directly in this richer latent space, the model might express more intermediate computations per step, achieving the same quality with fewer steps, or better quality with the same compute.

Recent work explores \emph{latent reasoning}, which aims to bypass discrete token generation. \citet{hao2024training} propose COCONUT that trains models to reason with continuous latent thoughts that are never decoded into language. \citet{cheng2024compressed} compress chain-of-thought into dense representations via knowledge distillation. Other approaches explore hybrid schemes that interleave latent and discrete tokens~\citep{su2025token, shen-etal-2025-codi, zhang-etal-2025-lightthinker}.

However, these methods face a key challenge: \emph{the lack of intermediate supervision}. Standard CoT training benefits from token-level cross-entropy loss at every reasoning step, providing dense gradient signal throughout the reasoning chain. When reasoning occurs in an unconstrained latent space, this supervision vanishes and the model must learn to produce useful intermediate representations without any direct feedback on what those representations should encode. This makes training unstable and prone to representational drift, particularly for long-horizon tasks where errors compound across many latent steps. As a result, prior latent reasoning methods have been demonstrated primarily on simple settings and often struggle to match the performance of explicit CoT on challenging benchmarks.

\begin{center}
\emph{Can we train the model to reason in a richer, superposed space while keeping intermediate supervision?}
\end{center}

\begin{figure*}[t]
    \centering
    \includegraphics[width=0.9\linewidth]{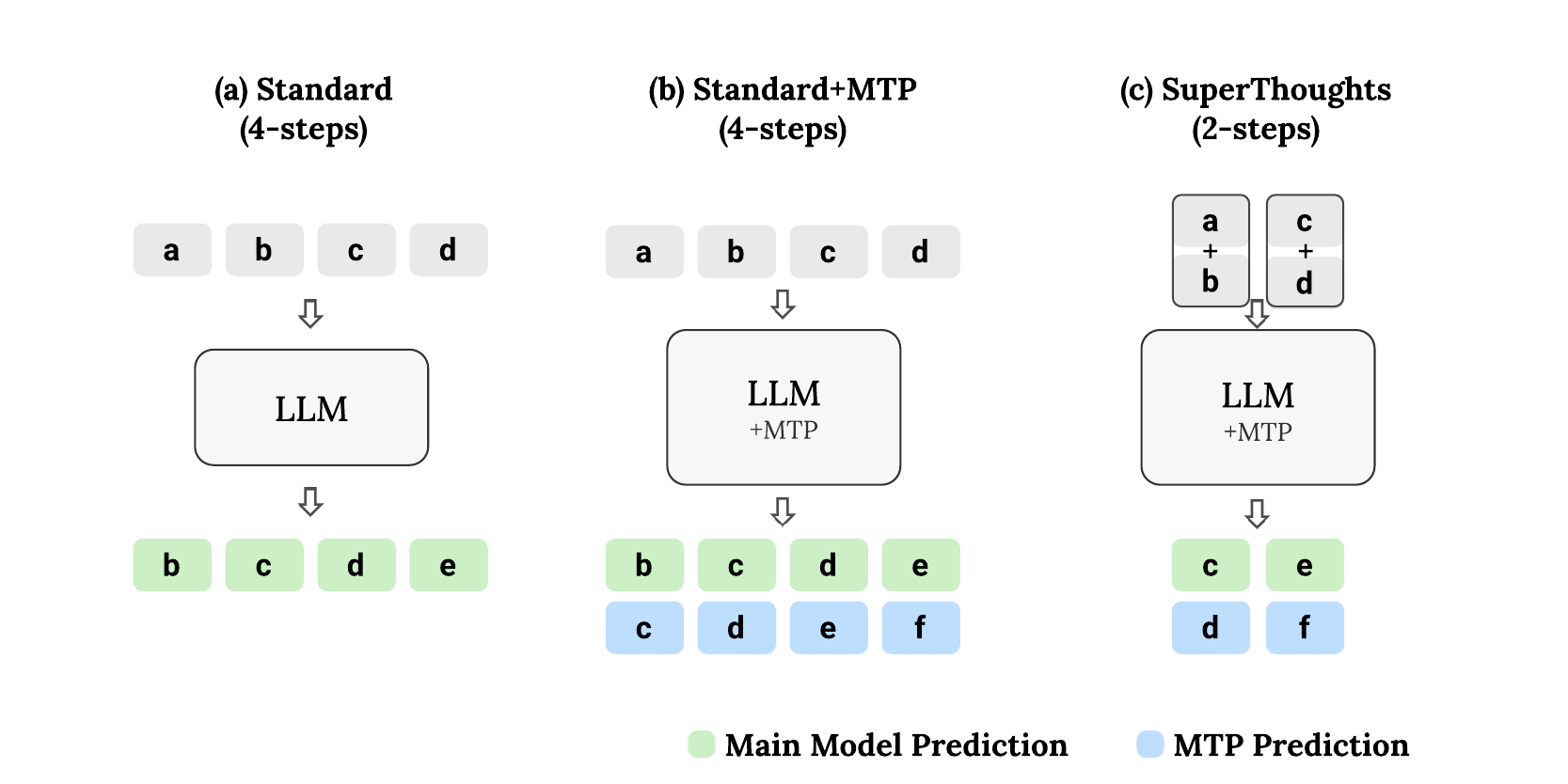}
    \caption{Comparison of three generation strategies for producing tokens ``b'' through ``e''. 
\textbf{(a) Standard:} Each forward pass consumes one token and predicts one token, requiring 4 steps. 
\textbf{(b) Standard + MTP:} A Multi-Token Prediction head predicts an additional token per step, but inputs remain single tokens, still requiring 4 steps. 
\textbf{(c) SuperThoughts:} Token pairs are fused into superposed embeddings as input, and two tokens are decoded per step via MTP, halving the required forward passes to 2 steps. 
Green denotes main model predictions; blue denotes MTP predictions.}
    \label{fig:overview}
\end{figure*}

In this work, we explore a natural first step toward this goal. We propose \textbf{SuperThoughts}, a framework that compresses \emph{pairs} of consecutive CoT tokens into single latent representations during reasoning. At each step, the model consumes a superposed embedding of two tokens and predicts two discrete tokens: one from the main model backbone and one from a lightweight Multi-Token Prediction (MTP) module~\citep{10.5555/3692070.3692699, liu2024deepseek}. This halves the number of forward passes required while maintaining token-level cross-entropy supervision throughout training.

Our main contributions are:
\begin{enumerate}
    \item We propose SuperThoughts, an architecture that compresses token pairs into single representations via a Compressor and decodes two tokens per step using a Main Module and an MTP Module.
    \item We develop a two-stage training protocol that first aligns the compressed latent space via distillation~\citep{berton2025compllmcompressionlongcontext}, then jointly trains all components end-to-end with discrete token supervision.
    \item We introduce a confidence-based adaptive inference mechanism that falls back to standard decoding when the MTP module is uncertain, trading throughput for accuracy on difficult reasoning steps.
    \item We evaluate on MATH500 \citep{hendrycksmath2021}, AMC23 \citep{MAA2023AMC}, OlympiadBench \citep{he2024olympiadbench} and GPQA-Diamond \citep{rein2024gpqa}, achieving $20$--$35\%$ CoT length reduction while maintaining accuracy within $1$--$2$ points of the baseline.
\end{enumerate}

\section{Related Works}
\paragraph{Latent Reasoning in LLMs.} When prompted with a question, LLMs can generate intermediate reasoning via discrete tokens before answering the question, and such reasoning process is termed chain-of-thought (CoT) \citep{wei2022chain}. Recently, several works focus on using CoT states beyond discrete tokens. \citet{hao2024training, yue2025hybrid, shen-etal-2025-codi} introduce methods that directly feed the last continuous hidden state as input embedding for the next step. However, these methods either require complicated training curriculum or only consider simple settings. \citet{giannou2025stoic,zhang-etal-2025-lightthinker,deng2025latentreasoningllmsvocabularyspace, shen2025hybridcot} generate first and then compress the newly generated tokens, but only save context length and involve attention mask manipulations that are not compatible with modern inference engines~\citep{10.1145/3600006.3613165, 10.5555/3737916.3739916}. \citet{cheng2024compressed, su2025token, tan2025thinksilentlythinkfast} trains the model to compress discrete CoT into latent tokens and during inference generate latent tokens directly. Several works explore composing multiple next token choices into a latent input token \citep{zhang2025soft,zhuang2025text,zhu2025reasoning,jain2025learningreasonmixturetokens,wu2025llmssinglethreadedreasonersdemystifying,tang2026multiplex,gozeten2026continuous}. \citet{peng2026efficient} pretrains LLMs with token superposition and yields pretraining time speedup. 

\paragraph{Compressed Input Context.} In addition to latent reasoning, there have been many works that compress more information into input embeddings. Prefix Tuning \citep{li-liang-2021-prefix} uses a learned soft embedding prefix to condition the LLM. Many works compress input context tokens to save context length  \citep{jiang2023llmlinguacompressingpromptsaccelerated, li-etal-2023-compressing, mu2023learning, berton2025compllmcompressionlongcontext, feldman2025simple}.

\paragraph{Reducing Discrete CoT Tokens.} Many methods have also produced shorter discrete CoT sequences through Reinforcement Learning~\citep{aggarwal2025l, shrivastava2025samplethinklessgroup} and fine-tuning~\citep{xia-etal-2025-tokenskip}. Notably, these discrete CoT tokens length reduction methods are orthogonal to SuperThoughts.

\paragraph{Variable Compute Per Token.} Recent work has explored adaptive compute allocation in language models by moving beyond uniform token-level processing, such as BLT \citep{pagnoni-etal-2025-byte} and H-Net \citep{hwang2025dynamicchunkingendtoendhierarchical} segmenting bytes into dynamically-sized patches, and DLCM \citep{qu2026dynamiclargeconceptmodels} learning variable-length semantic concepts on top of tokens. \citet{liu2025hamburgeracceleratingllminference} propose HAMBURGER, which similarly fuses multiple tokens into a single input embedding via a compositional embedder and decodes several tokens per forward through a micro-step decoder.

\paragraph{Multi-token Prediction.} Traditionally, LLMs are trained with next-token prediction loss where the model is provided with a prefix and asked to predict the next token that follows the prefix \citep{radford2019language}. \citet{pmlr-v235-bachmann24a} argue that teacher-forcing in next-token prediction results in inaccurate next-token predictor and proposes a solution that learns to predict multiple tokens. \citet{10.5555/3692070.3692699} pre-train LLMs from scratch that predicts multiple future tokens at once using multiple output heads and show that multi-token prediction (MTP) is better than next-token prediction (NTP) on larger models. DeepSeek-V3 \citep{liu2024deepseek} also train the model with MTP objective but use a lightweight MTP module instead of an independent output head. \citet{ahn2025efficient} propose joint multi-token prediction (JTP) by employing a representation bottleneck that encourages the model to encode richer information in the output hidden state.

Despite MTP predicting multiple tokens at once, at inference time current MTP architectures can only utilize the extra tokens for self-speculative decoding \citep{liu2024deepseek, 10.5555/3692070.3692699, 10.5555/3692070.3692273}, since the main model still needs to populate KV entries for tokens the MTP modules generate. Critically, this does not reduce the total FLOPs at inference time. The main model must still perform a full forward pass over every accepted token, meaning self-speculative decoding with MTP targets latency reduction under low GPU utilization rather than computational efficiency.

\paragraph{Scaling Test-Time Compute.}
Recent scaling laws suggest that optimizing test-time compute can outperform simply increasing parameter counts  \citep{snell2024scaling}. Leading reasoning models, such as OpenAI o1 \citep{jaech2024openai} and DeepSeek-R1 \citep{guo2025deepseek}, utilize Reinforcement Learning extended CoT sequences.

\section{Methods}
\begin{figure*}[t]
    \centering
    \includegraphics[width=0.65\linewidth]{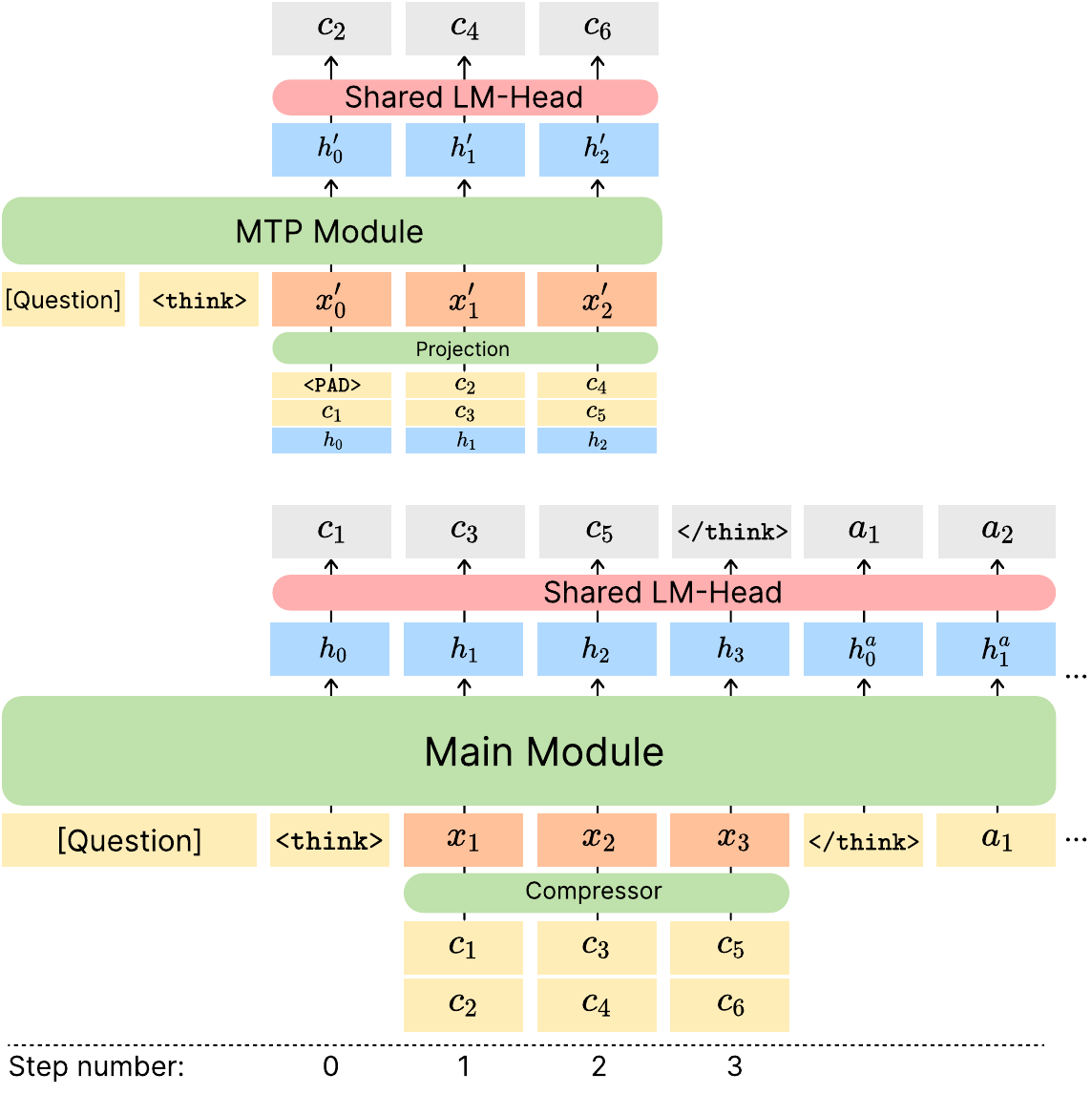}
    \caption{\textbf{Overview of SuperThoughts architecture.} At each step $i$, the Compressor encodes a CoT token pair $(c_{2i-1}, c_{2i})$ into a single latent vector $\vx_i$ via a learned $2H \to H$ compressor, where $H$ is the dimension of a single token embedding. The Main module processes $\vx_i$ to produce hidden state $\vh_i$ and predicts the next odd-indexed token $c_{2i+1}$. The MTP module then receives a projection $\vx'_i$ that combines three inputs -- the previous even token, the just-predicted odd token from Main, and the Main hidden state -- via a learned $3H \to H$ projection, and predicts the corresponding even-indexed token $c_{2i+2}$. Both modules share the same output LM head. This design enables the model to consume two tokens and generate two tokens per step.}
    \label{fig:superthoughts}
\end{figure*}
Standard Chain-of-Thought (CoT) reasoning generates a sequence of discrete tokens $c_{1:L} = (c_1, \dots, c_L)$ autoregressively, requiring $L$ forward passes to produce a reasoning chain of length $L$. We introduce \textbf{SuperThoughts}, a framework that halves this computational cost by processing and generating tokens in pairs. At each reasoning step, the model consumes two tokens and predicts two tokens, reducing the number of forward passes from $L$ to $L/2$ while preserving discrete token supervision.

In this section, we detail: \textbf{(1) Architecture~(\Cref{sec:arch}):} The three components of our model that includes a Compressor, a Main module, and a lightweight Multi-Token Prediction (MTP) Module; \textbf{(2) Training~(\Cref{sec:training}):} A two-stage protocol that first aligns the compressed latent space via distillation, then jointly trains all components; and \textbf{(3) Adaptive Inference~(\Cref{sec:inference}):} A decoding algorithm that dynamically falls back to standard single-token generation when model confidence is low.

\subsection{SuperThoughts Architecture}\label{sec:arch}

Our model processes reasoning chains by operating on superposed token pairs during the thinking process rather than individual CoT tokens. We structure each example as a sequence
\[
\underbrace{q_{1:L_q}\ \texttt{<think>}}_{\text{prompt tokens}}\ \underbrace{c_{1:L_c}\ \texttt{</think>}\ a_{1:L_a}}_{\text{response tokens}},
\]
where $q_{1:L_q}$ denotes question tokens, $c_{1:L_c}$ denotes CoT tokens, and $a_{1:L_a}$ denotes answer tokens. The sequence includes special delimiter tokens \texttt{<think>} and \texttt{</think>}, with \texttt{<think>} appended to the prompt to initiate reasoning. We reorganize the CoT sequence $c_{1:L_c}$ into a sequence of pairs, reducing the effective reasoning length from $L_c$ to $S=L_c/2$ steps, where we assume $L_c$ is even and $S$ represents the number of superposition steps; if $L_c$ is odd, we pad with a special token to maintain the pair structure.

Our model consists of three components: (1) Compressor, (2) Main module and (3) Multi-Token Prediction (MTP) Module. At CoT phase, for each step $i$, the Compressor encodes a token pair $(c_{2i-1}, c_{2i})$ into a single latent vector, from which the Main module predicts the next token $c_{2i+1}$ and the MTP module predicts the next-next token $c_{2i+2}$.

\paragraph{Compressor.}
Let $\texttt{Emb}(\cdot) \in \mathbb{R}^H$ denote the token embedding function. For each step $i=1,\ldots,S$, the compressor $\texttt{Comp}(\cdot)$ maps the pair $(\texttt{Emb}(c_{2i-1}), \texttt{Emb}(c_{2i}))$ into a single compressed vector $\vx_i \in \mathbb{R}^H$. We explore two implementations for the compressor: either a \textit{Linear Projection}, where we concatenate the token embeddings and project them using a learnable matrix $P \in \mathbb{R}^{H \times 2H}$:
\[ \vx_i = P\begin{bmatrix}\texttt{Emb}(c_{2i-1})\\ \texttt{Emb}(c_{2i})\end{bmatrix}, \]
or a \textit{Transformer Block}, where we process the pair using a small Transformer layer and extract the hidden state corresponding to the second token (denoted by the subscript $2$):
\[ \vx_i = \texttt{TF}\big(\texttt{Emb}(c_{2i-1}), \texttt{Emb}(c_{2i})\big)_2, \]
where $[\cdot, \cdot]$ denotes sequence concatenation. These compressed vectors $\vx_{1:S}$ serve as the inputs to the Main module.

\paragraph{Main module.}
The Main module (base LLM) is the primary reasoning backbone, responsible for evolving the latent reasoning state and predicting \textit{odd-indexed} tokens. At each time step, it takes as input the compressed representation of two tokens, and outputs a latent reasoning state, which is fed to a language modeling head to predict the next token. In more detail, step $i=0$, the Main module takes in \texttt{<think>}, produces $\vh_0$ and predicts $c_1$; at steps $i\geq 1$, it takes in $\vx_i$, produces the hidden state $\vh_i$ and predicts $c_{2i+1}$. At the final step $i=S$, it predicts the closing delimiter \texttt{</think>} from $\vh_S$. The Main and MTP modules share the same output language modeling head.

\paragraph{MTP module.}
The MTP module is responsible for predicting \textit{even-indexed} tokens. At each step, it takes as input the hidden representation output from Main module, some of the past tokens, and predicts next to next token. In more detail, at step $i$, it takes in the previous even token $c_{2i}$ (with $c_0=\texttt{<PAD>}$), the current odd token $c_{2i+1}$ just predicted by Main and the hidden state $\vh_i$ from Main. These are compressed into $\vx'_i\in\mathbb{R}^H$ via a learnable projection $P'\in\mathbb{R}^{H\times 3H}$:
\[
\vx'_i = P'\begin{bmatrix}
\texttt{RMSNorm}(\texttt{Emb}(c_{2i}))\\
\texttt{RMSNorm}(\texttt{Emb}(c_{2i+1}))\\
\texttt{RMSNorm}(\vh_{i})
\end{bmatrix}.
\]
A one-layer transformer then produces $\vh'_i$ and predicts $c_{2i+2}$.

\subsection{Training Strategy}\label{sec:training}

Training a model to reason in latent space can be unstable if the compressed representations drift significantly from the pre-trained language manifold. To mitigate this, we employ a two-stage training protocol: first, we warm-start the compressor via knowledge distillation to align the latent space; second, we jointly train the entire model using standard cross-entropy loss.

\paragraph{Stage 1: Training Compressor module via latent distillation.}
\begin{figure*}[t!]
    \centering
    \includegraphics[width=0.9\linewidth]{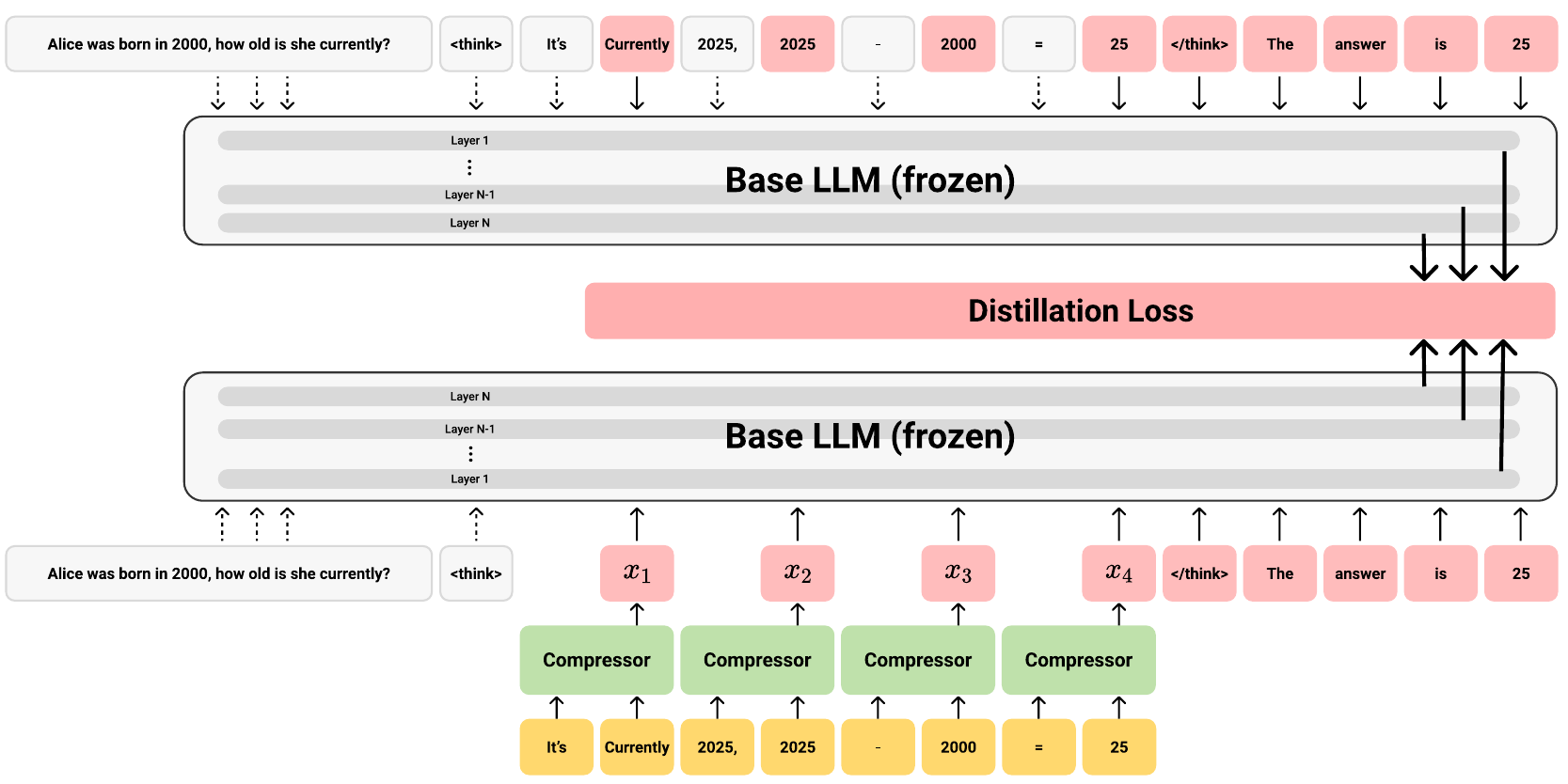}
    \caption{\textbf{Compressor training via latent distillation.} \textbf{Top (Teacher):} The frozen Base LLM processes the full discrete token sequence. \textbf{Bottom (Student):} The same frozen Base LLM receives compressed representations $\vx_i$ from the Compressor, which fuses each CoT token pair (e.g., ``It's'' + ``Currently'' $\to$ $\vx_1$). The Compressor is trained to minimize the Smooth-$L_1$ distance between teacher and student hidden states across all layers at corresponding positions. Positions used to compute the distillation loss are marked in red.}
    \vspace{-0.05pt}
    \label{fig:distill-loss}
\end{figure*}
Before end-to-end training, we train the Compressor module via distillation, following \citet{berton2025compllmcompressionlongcontext}. Consider the Main module processing the discrete sequence $[c_1,c_2,\ldots,c_{2S}]$ one token at a time (the \textit{teacher}), versus the Main module processing compressed pairs $[\vx_1, \ldots, \vx_S]$ where $\vx_i=\texttt{Comp}(c_{2i-1}, c_{2i})$ (the \textit{student}). We train the Compressor so that the student's hidden state after $\vx_i$ matches the teacher's hidden state (layer-wise) after $c_{2i}$. In effect, the model should produce the same hidden states whether processing tokens discretely or in compressed form.

We define the set of distillation targets $D$ as pairs of corresponding positions in the teacher (uncompressed) and student (compressed) sequences. Following \citet{berton2025compllmcompressionlongcontext}, we include all answer token positions. Crucially, we extend their approach by also including all \textit{even} CoT token positions, enforcing alignment within the reasoning chain itself and not just the final answer.

The loss is computed as the Smooth-$L_1$ distance between the teacher's hidden states $H^{(\ell)}$ and the student's hidden states $\tilde{H}^{(\ell)}$ across all layers $\ell$ and target pairs $(t, t') \in D$:
\[
\mathcal{L}_{\text{distill}} = \sum_{\ell} \frac{1}{\sigma^{(\ell)} |D|} \sum_{(t,t') \in D} \operatorname{SmoothL1}_{\beta}\bigl({H}^{(\ell)}_{t},\, \tilde{H}^{(\ell)}_{t'}\bigr),
\]
where $\sigma^{(\ell)} = \operatorname{Std}(H^{(\ell)}_{D})$ is layer-wise normalization and the Smooth-$L_1$ distance $\operatorname{SmoothL1}_{\beta}(u,v)$ is defined as $\frac{1}{d}\sum^d_{i = 0}\operatorname{SmoothL1}_{\beta}(u,v)_i$ with
\[
\operatorname{SmoothL1}_{\beta}(u, v)_i =
\begin{cases}
\dfrac{1}{2}\dfrac{(u_i - v_i)^2}{\beta}, & |u_i - v_i| < \beta, \\[6pt]
|u_i - v_i| - \dfrac{\beta}{2}, & \text{otherwise}.
\end{cases}
\]

\paragraph{Stage 2: Joint training with cross-entropy loss.}
Once the compressor is aligned, we train the full model (Main + MTP + Compressor) end-to-end. This essentially minimizes the cross entropy loss for all tokens predicted (including those coming from the Main module and the MTP module). Let $\ell(y\mid\vh):=\mathrm{CE}\!\bigl(y,\mathrm{head}(\vh)\bigr)$ denote token-level cross-entropy (CE) loss. Define the Main targets $y^{\mathrm{main}}_i=c_{2i+1}$ for $i=0, 1,\ldots,S-1$, and $y^{\mathrm{main}}_S=\texttt{</think>}$.
The CoT losses are
\begin{align*}
\mathcal{L}^{\mathrm{CoT}}_{\mathrm{NTP}}
&= \frac{1}{S+1}
\sum_{i=0}^{S}\ell\!\left(y^{\mathrm{main}}_i \mid \vh_i\right), \\
\mathcal{L}^\mathrm{CoT}_{\mathrm{MTP}}
&= \frac{1}{S}
\sum_{i=0}^{S-1}\ell\!\left(c_{2i+2} \mid \vh'_i\right).
\end{align*}
Let $\mathcal{L}_{\mathrm{answer}}$ denote the standard next-token CE loss on the answer tokens. We define
\[
\mathcal{L}_{\mathrm{NTP}} := \mathcal{L}_{\mathrm{answer}} + \mathcal{L}^{\mathrm{CoT}}_{\mathrm{NTP}}, \mathcal{L}_{\mathrm{Training}} = \mathcal{L}_{\mathrm{NTP}} + \lambda\,\mathcal{L}^\mathrm{CoT}_{\mathrm{MTP}}.
\]

Since we train the MTP module from scratch, we first freeze the Main and Compressor module to train the MTP module, after which we unfreeze all modules and jointly train them.

\subsection{Confidence-based Adaptive Inference}\label{sec:inference}
While our model can process two CoT tokens within one step with superposed tokens, this sometimes can degrade the performance. For example, if the model needs to decode two ``hard'' tokens, a single superposition step may lack sufficient computational capacity to predict both correctly. Ideally, the model should allocate more compute to ``hard tokens'' via discrete reasoning, while processing ``easy tokens'' efficiently with superposition.

We implement this by looking at the confidence of the MTP module. At each step $i$, after the Main module predicts the odd token $c_{2i+1}$, the MTP module produces $\vh'_i$ that will be used to predict $c_{2i+2}$. Let 
\begin{align*}
p_{i}^{\text{MTP}} = \max_{j\in V} \text{softmax}(\text{head}(\vh'_i))_j
\end{align*}
be the maximum probability of the MTP prediction at step $i$ and $\tau$ be a threshold. If $p_{i}^{\text{MTP}} < \tau$, this means the MTP module is not confident about the prediction. In this case, we reject the MTP prediction. On the next step $i+1$, instead of feeding a compressed pair to the Main module, we input $\texttt{Emb}(c_{2i+1})$ directly and re-predict $c_{2i+2}$ using the more powerful Main module, after which MTP tries to predict $c_{2i+3}$ followed by the same acceptance check. This fallback mechanism allows the model to self-regulate its speed, processing two tokens for easy text while slowing down for difficult reasoning steps. We analyze the inference cost in Appendix \ref{app:inf_cost}.

\section{Experiments}
\begin{table*}[t!]
    \caption{Accuracy and average correct CoT length of Qwen-2.5-Math-Instruct-1.5B/7B models on three benchmarks. We trained two variants of SuperThoughts model, one with a projection matrix as the Compressor and another with a 1-layer Transformer as the Compressor. The baseline CoT is trained on the same dataset as we train SuperThoughts.}
    \label{tab:main_result}
    \centering
    \resizebox{\textwidth}{!}{%
    \begin{tabular}{ll|cc|cc|cc}
        \toprule
        & & \multicolumn{2}{c|}{MATH500} & \multicolumn{2}{c|}{OlympiadBench} & \multicolumn{2}{c}{AMC23} \\
        & & Acc (\%) & CoT Len. & Acc (\%) & CoT Len. & Acc (\%) & CoT Len. \\
        \midrule 
        \multirow{11}{*}[-1.5ex]{\rotatebox{90}{Qwen2.5-Math-1.5B-Instruct}} & Standard CoT & 72.4 & 494.1 & 34.72 & 620.5 & 59.53 & 660.5 \\
        \cmidrule(lr){2-8}
        & \multicolumn{7}{c}{\textbf{SuperThoughts \emph{(Projection Compressor)}}} \\
        \cmidrule(lr){2-8}
        & \emph{w/o adaptive} & 55.6 & 229.6 & 20.03 & 286.9 & 38.28 & 351.0 \\
        & \emph{w/ adaptive, $\tau=0.999$} & 73.0 & 314.8 & 33.83 & 436.3 & 57.97 & 471.6 \\
        & \emph{w/ adaptive, $\tau=0.9999$} & 72.0 & 355.5 & 33.09 & 463.0 & 57.03 & 531.1 \\
        & \emph{w/ adaptive, $\tau=0.99999$} & 71.4 & 408.5 & 34.42 & 537.3 & 58.13 & 590.9 \\
        \cmidrule(lr){2-8}
        & \multicolumn{7}{c}{\textbf{SuperThoughts \emph{(Transformer Compressor)}}} \\
        \cmidrule(lr){2-8}
        & \emph{w/o adaptive} & 55.8 & 224.9 & 21.01 & 289.4 & 40.78 & 348.7 \\
        & \emph{w/ adaptive, $\tau=0.999$} & 72.2 & 319.8 & 32.49 & 423.7 & 56.41 & 459.3 \\
        & \emph{w/ adaptive, $\tau=0.9999$} & 71.0 & 354.2 & 33.68 & 463.6 & 57.66 & 524.9 \\
        & \emph{w/ adaptive, $\tau=0.99999$} & 72.0 & 406.7 & 34.13 & 530.7 & 58.44 & 562.7 \\
        \midrule \\
        \midrule 
       \multirow{11}{*}[-2ex]{\rotatebox{90}{Qwen2.5-Math-7B-Instruct}} & Standard CoT  & 83.0 & 538.6 & 36.35 & 654.6 & 69.22 & 749.4 \\
        \cmidrule(lr){2-8}
        & \multicolumn{7}{c}{\textbf{SuperThoughts \emph{(Projection Compressor)}}} \\
        \cmidrule(lr){2-8}
        & \emph{w/o adaptive} & 72.4 & 252.1 & 30.71 & 339.5 & 57.11 & 359.2 \\
        & \emph{w/ adaptive, $\tau=0.999$} & 81.0 & 334.7 & 35.31 & 415.6 & 66.72 & 467.2 \\
        & \emph{w/ adaptive, $\tau=0.9999$} & 80.8 & 357.3 & 35.46 & 456.8 & 67.89 & 509.9 \\
        & \emph{w/ adaptive, $\tau=0.99999$} & 82.4 & 404.2 & 38.13 & 553.0 & 67.11 & 583.1 \\
        \cmidrule(lr){2-8}
        & \multicolumn{7}{c}{\textbf{SuperThoughts \emph{(Transformer Compressor)}}} \\
        \cmidrule(lr){2-8}
        & \emph{w/o adaptive} & 72.0 & 251.1 & 27.89 & 336.5 & 57.97 & 363.0 \\
        & \emph{w/ adaptive, $\tau=0.999$} & 80.4 & 323.6 & 37.24 & 425.6 & 67.19 & 467.1 \\
        & \emph{w/ adaptive, $\tau=0.9999$} & 80.8 & 360.3 & 35.46 & 442.2 & 67.66 & 514.8 \\
        & \emph{w/ adaptive, $\tau=0.99999$} & 82.0 & 411.2 & 37.24 & 541.1 & 66.02 & 539.3 \\
        \bottomrule
    \end{tabular}
    }
\end{table*}

\subsection{Experimental Setup}
In this section, we introduce our experiment to train a model that reasons in superposition. We use Qwen2.5-Math-1.5B-Instruct and Qwen2.5-Math-7B-Instruct as the model we start from and post-train it to reason in superposition. The Main module is initialized the same as Qwen2.5 and the MTP module is initialized using the weights of the last layer. The projection matrices are initialized as a map that averages the input embeddings.

To train the model, we curate a synthetic reasoning dataset. We collect questions from \citet{albalak2025bigmathlargescalehighqualitymath} and \citet{moshkov2025aimo2}. For each question, we let Qwen2.5 to generate 10 responses with CoT. We filter out incorrect responses and further filter down to $\sim$1.5M responses. For each sample, we consider the whole response as the CoT part and the ``\textbackslash boxed\{...\}'' part as the answer part. The dataset preparation process is detailed in Appendix \ref{alg:dataset-prep}.

The Main module is initialized the same as the pretrained Qwen2.5 model. In stage 1 training, we choose a projection matrix or a 1-layer transformer as two variants of the Compressor module and uses learning rate $1\times 10^{-4}$.

In stage 2 training, we first freeze the Compressor and the Main module and only train MTP with learning rate $5\times 10^{-4}$. Then for the joint training, we use learning rate $1\times 10^{-5}$ and train for 2 epochs. For model trained without using adaptive inference, we use $\lambda=1.0$; for model trained to perform adaptive inference, we use $\lambda = 0.02$. This is because since uncertain MTP predictions can be rejected and re-predicted by the Main module in the subsequent step, we prioritize the Main module's accuracy and thus do not need to optimize the MTP loss as aggressively. We train Qwen2.5 on the same dataset as a baseline and discuss our baseline choice in Appendix \ref{app:baseline}.

\subsection{Results}
\begin{highlight}
    SuperThoughts saves CoT length by 20-30\% with confidence-based adaptive inference while only having a small accuracy drops (1-2 points on most tasks).
\end{highlight}
Table~\ref{tab:main_result} presents our main results on three mathematical reasoning benchmarks: MATH500, OlympiadBench, and AMC23. We evaluate both compressor variants (Projection and Transformer) across different inference settings.

\paragraph{Uniform superposition reduces length but hurts accuracy.}
Without adaptive inference, SuperThoughts reduces CoT length by approximately half but incurs substantial accuracy drops. On \textbf{1.5B} (Projection), CoT length decreases by $47$--$54\%$ across benchmarks, but accuracy drops by $14.7$--$21.3$ points. On \textbf{7B} (Projection), compression rates are similar ($48$--$53\%$), but accuracy drops are notably smaller at $5.6$--$12.1$ points. This \emph{scale effect} suggests that larger models better tolerate aggressive token compression. Both compressor variants exhibit similar behavior, indicating the bottleneck is per-step compute capacity rather than the compression mechanism.

\paragraph{Adaptive inference recovers accuracy.}
Confidence-based adaptive inference substantially closes the accuracy gap while retaining meaningful CoT length reductions. On \textbf{1.5B} (Projection) with $\tau=0.999$, MATH500 accuracy matches the baseline ($73.0\%$ vs.\ $72.4\%$) with a $36\%$ CoT reduction. OlympiadBench and AMC23 remain within $0.9$--$1.6$ points of baseline while achieving $29$--$30\%$ reductions. On \textbf{7B} (Projection), $\tau=0.9999$ offers a balanced trade-off: $30$--$34\%$ CoT reduction with accuracy within $0.9$--$2.2$ points across all benchmarks.

Notably, while larger $\tau$ sometimes achieves best accuracy, increasing $\tau$ \emph{does not always improve accuracy}. On \textbf{1.5B} (Projection) MATH500, accuracy is the highest at $\tau=0.999$. Similarly, on \textbf{7B} (Projection) AMC23, the best adaptive accuracy occurs at $\tau=0.9999$. We hypothesize this reflects noise or that $\tau=0.999$ already provides a sufficiently high confidence threshold.

\paragraph{Compressor comparison.}
The Projection and Transformer compressors perform similarly. Without adaptive inference, Projection shows a slight edge (e.g., $30.7\%$ vs.\ $27.9\%$ on 7B OlympiadBench), but this gap disappears with adaptive decoding. Given its simplicity and lower computational cost, the linear projection is the preferred choice.

\paragraph{Comparison with HAMburger.} We train HAMburger \citep{liu2025hamburgeracceleratingllminference} using the same data on Qwen2.5-1.5B and compare against SuperThoughts. We choose confidence $\in\{0.93, 0.95, 0.99, 0.995, 0.999, 0.9995, 0.9999, 1.0\}$ and for SuperThoughts we choose $\tau\in\{0.99, 0.993, 0.995, 0.999, 0.9999, 0.99999\}$. For each configuration we record the CoT length and the accuracy in Figure \ref{fig:hamburger_comp}. SuperThoughts achieves higher accuracy at every compression level, and shorter CoT at every accuracy level, than HAMburger.

\begin{figure*}[t]
    \centering
    \includegraphics[width=\linewidth]{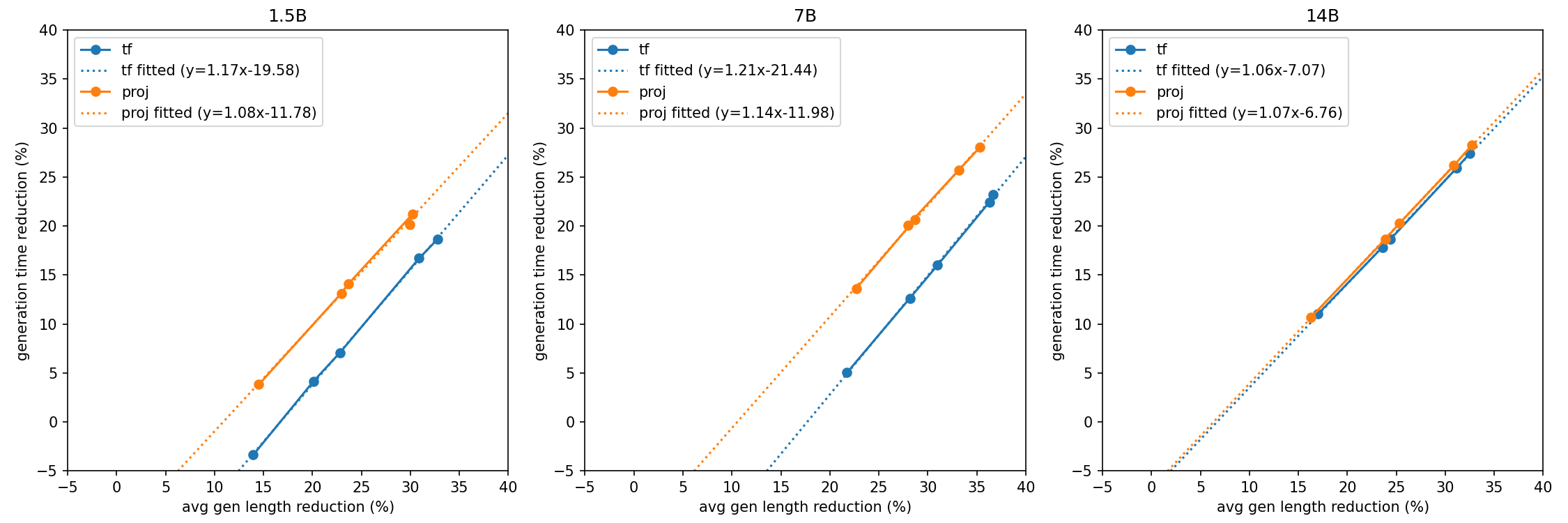}
    \caption{Generation length reduction vs. wall-clock time inference speedup on 1.5B, 7B and 14B models.}
    \label{fig:wallclock}
\end{figure*}

\paragraph{Beyond mathematical reasoning.}
\begin{table*}[t!]
    \caption{Accuracy and average correct CoT length of Qwen-2.5-Instruct-14B trained models\protect\footnotemark on four benchmarks.}
    \label{tab:14b_result}
    \centering
    \resizebox{\textwidth}{!}{%
    \begin{tabular}{l|cc|cc|cc|cc}
        \toprule
        & \multicolumn{2}{c|}{MATH500} & \multicolumn{2}{c|}{OlympiadBench} & \multicolumn{2}{c}{AMC23} & \multicolumn{2}{c}{GPQA-Diamond}\\
        & Acc (\%) & CoT Len. & Acc (\%) & CoT Len. & Acc (\%) & CoT Len. & Acc (\%) & CoT Len. \\
        \midrule
        Standard CoT & 78.0 & 506.1 & 38.0 & 651.1 & 65.5 & 723.6 & 41.5 & 675.3 \\
        \midrule
        \multicolumn{9}{c}{\textbf{SuperThoughts \emph{(Projection Compressor)}}} \\
        \midrule
        \emph{w/ adaptive, $\tau=0.999$} & 77.4 & 315.7 & 35.5 & 441.9 & 63.0 & 465.3 & 41.5 & 510.2 \\
        \emph{w/ adaptive, $\tau=0.9999$} & 75.8 & 354.5 & 35.2 & 481.8 & 62.0 & 504.9 & 42.9 & 554.0 \\
        \emph{w/ adaptive, $\tau=0.99999$} & 77.6 & 408.0 & 36.6 & 519.4 & 66.4 & 581.2 & 40.0 & 615.2 \\
        \midrule
        \multicolumn{9}{c}{\textbf{SuperThoughts \emph{(Transformer Compressor)}}} \\
        \midrule
        \emph{w/ adaptive, $\tau=0.999$} & 77.2 & 315.1 & 36.7 & 433.7 & 63.1 & 462.5 & 39.1 & 502.1 \\
        \emph{w/ adaptive, $\tau=0.9999$} & 78.6 & 356.9 & 38.3 & 486.1 & 63.9 & 507.1 & 43.0 & 575.3 \\
        \emph{w/ adaptive, $\tau=0.99999$} & 77.0 & 398.1 & 37.2 & 548.0 & 63.3 & 562.5 & 41.2 & 607.7 \\
        \bottomrule
    \end{tabular}
    }
\end{table*}
To test whether \textbf{SuperThoughts} applies to reasoning capabilities beyond Math, we add additional science questions from \citet{guha2025openthoughtsdatarecipesreasoning} and train Qwen2.5-14B-Instruct (a non-Math Instruct model) following the same training paradigm. Table \ref{tab:14b_result} shows that \textbf{SuperThoughts} works on domains other than Math. \footnotetext{Note that this is a non-Math model so the accuracies on Math benchmarks are lower than the 7B-Math model in Table \ref{tab:main_result}; for the Projection compressor we add an additional RMSNorm after the compressor.}
\begin{highlight}
    The gap between theoretical speedup (CoT length reduction) and the actual speedup (wall-clock time reduction) is smaller as the model gets larger.
\end{highlight}
\paragraph{Inference wall-clock time analysis.}
We run additional experiment to measure the theoretical speedup (generation length reduction) vs actual speedup (generation time reduction). We implement SuperThoughts using nano-vLLM\citep{nanovllm2025} for fast inference. We run MATH500 (500 questions) on 1.5B, 7B and 14B model, and for each model we choose $\tau\in\{0.999, 0.9995, 0.9999, 0.99995, 0.99999\}$. For each generation configuration, we record generation length reduction $R_{\text{len}}$ and wallclock generation time reduction $R_{\text{time}}$: $$R_{\text{len}}=\frac{L_{\text{baseline}}-L_{\text{SuperThoughts}}}{L_{\text{baseline}}},$$
$$R_{\text{time}}=\frac{T_{\text{baseline}}-T_{\text{SuperThoughts}}}{T_{\text{baseline}}},$$ where $L_{\text{baseline}}$ is the baseline average generation length, $L_{\text{SuperThoughts}}$ is the SuperThoughts average generation length, $T_{\text{baseline}}$ is the baseline generation time and $T_{\text{SuperThoughts}}$ is the SuperThoughts generation time. We plot $R_{\text{len}}$ versus $R_{\text{time}}$ in Figure \ref{fig:wallclock}. We can see that the additional overhead from SuperThoughts (compressor, MTP and adaptive fallback) has less influence as the model gets larger. For example, with projection compressor, on 1.5B model, 30.25\% reduction in generation length gives 21.25\% reduction in generation time; on 7B model, 33.17\% reduction in generation length gives 25.72\% reduction in generation time; on 14B model, 32.75\% reduction in generation length gives 28.3\% reduction in generation time. This is because while the compressor, MTP, and adaptive fallback adds additional overhead (e.g., additional kernel launches and other CPU activities), such overhead takes a smaller percentage of the total generation time as the model gets larger.

\section{Discussion}
Our results show that SuperThoughts successfully compresses CoT reasoning while preserving accuracy: with adaptive inference, we achieve $20-35\%$ length reduction within  points below baseline across all benchmarks.

The adaptive mechanism also speaks to a broader question: \emph{how should models allocate compute across reasoning steps?} Standard decoding spends the same compute on every token, yet intuitively some steps can be harder than other steps. Our confidence-based adaptive mechanism can be viewed as a simple compute scheduler: superpose when the MTP module is confident, fall back to discrete tokens when it is not fully confident.

We also observe that larger models tolerate aggressive compression better. Under uniform superposition (no adaptive fallback), the 7B model drops $5-12$ points while the 1.5B model drops $14-21$ points. This suggests larger models have enough spare capacity per step to fit two tokens more reliably. If the trend holds at greater scale, even larger models might tolerate superposing three or more tokens, or need fewer fallbacks to discrete decoding.

Finally, we train the MTP module from scratch because Qwen2.5 lacks a native MTP module. Recent models like Qwen3-Next \citep{qwen3technicalreport} and and MiMo \citep{xiaomi2025mimo} are pre-trained with native MTP modules. Starting from a native MTP module would simplify training and likely improve results, since the MTP module is already aligned with the backbone. As native MTP becomes standard, SuperThoughts becomes easier to apply.

\bibliography{preprint}
\bibliographystyle{preprint}

\clearpage
\appendix

\section{Discussion on Baseline}
\label{app:baseline}
We report the standard discrete CoT as our baseline choice. In this section we discuss other related methods and why we didn't choose them as baselines.
\paragraph{Methods that reduce the discrete tokens.} Methods like TokenSkip \citep{xia-etal-2025-tokenskip} cuts the number of discrete CoT tokens by finetuning the model on a more concise reasoning trace. TokenSkip operates entirely within the discrete token space -- it reduces the number of tokens generated but still decodes one token per forward pass. \textbf{SuperThoughts}, by contrast, reduces the number of forward passes themselves by decoding multiple tokens per step, targeting a different axis of efficiency. The two approaches are complementary and could in principle be combined.

\paragraph{MTP for self-speculative decoding.} MTP module can be used as a draft model to speculatively predict future tokens and verify with the target model. However, speculative decoding does not reduce total FLOPs -- the main model must still perform full forward passes to populate KV cache entries for every accepted token. It can speed up inference in low-utilization regimes with small batch sizes by increasing GPU utilization, but when serving large batches at high throughput, the GPU is already compute-bound and no speed-up is achieved. On the other hand, \textbf{SuperThoughts} reduces total FLOPs by adaptively generating multiple tokens per forward pass, in contrast to standard autoregressive decoding which produces a single token per step. Therefore, speculative decoding is not a suitable baseline for \textbf{SuperThoughts}, as it targets a fundamentally different bottleneck -- latency under low utilization -- rather than reducing total FLOPs, which is the focus of our work.

\paragraph{Latent reasoning.} Latent reasoning methods like COCONUT \citep{hao2024training}, CODI \citep{shen-etal-2025-codi} and CoLaR \citep{tan2025thinksilentlythinkfast} directly use the latent from the last layer as the input to the next reasoning step. However, these methods are designed for tasks and training regimes that involve short reasoning sequences (e.g., 20–60 tokens), and we find that they do not translate effectively to our settin. For example we tried to train CODI in our setting and the trained model can hardly produce a correct answer (accuracy ~0.1\%). We conduct experiment with Soft Thinking \citep{zhang2025soft} that aggregates token embeddings from multiple candidates within a single step. However, Table \ref{tab:soft_thinking} shows that Soft Thinking is not significantly better than Standard CoT (except 1.5B on AMC23) and the CoT length is longer.

\begin{table*}[h!]
    \caption{Accuracy and average correct CoT length of Qwen-2.5-Math-Instruct-1.5B/7B models on three benchmarks, comparing the Standard CoT baseline against Soft Thinking.}
    \label{tab:soft_thinking}
    \centering
    \resizebox{\textwidth}{!}{%
    \begin{tabular}{ll|cc|cc|cc}
        \toprule
        & & \multicolumn{2}{c|}{MATH500} & \multicolumn{2}{c|}{OlympiadBench} & \multicolumn{2}{c}{AMC23} \\
        & & Acc (\%) & CoT Len. & Acc (\%) & CoT Len. & Acc (\%) & CoT Len. \\
        \midrule 
        \multirow{2}{*}{1.5B} & Standard CoT & 72.4 & 494.1 & 34.72 & 620.5 & 59.53 & 660.5 \\
        & Soft Thinking & 72.8 & 570.4 & 33.68 & 780.6 & 66.84 & 838.6 \\
        \midrule
        \multirow{2}{*}{7B} & Standard CoT  & 83.0 & 538.6 & 36.35 & 654.6 & 69.22 & 749.4 \\
        & Soft Thinking & 82.0 & 622.0 & 37.39 & 868.7 & 64.61 & 812.7 \\
        \bottomrule
    \end{tabular}
    }
\end{table*}

\section{Inference Cost Analysis}
\label{app:inf_cost}
In this subsection we analyze how much compute can \textbf{SuperThoughts} save compared to standard discrete CoT decoding. For simplicity we assume batch size $B=1$; the analysis extends directly to $B>1$ since $B$ cancels in the ratio $C_{\text{ST}}/C_{\text{base}}$, and \textbf{SuperThoughts} supports batched generation as token superposition operates independently within each sequence. Let $L_q$ be the question (prompt) length, $L_c$ be the CoT length and $L_a$ be the response length. The inference cost can be split into prefilling cost and decoding cost, while at prefilling stage the model process the question part one time and at the decoding stage the model generates CoT and answer autoregressively. For reasoning intensive tasks like Math, usually $L_c + L_a \gg L_q$, meaning that the decoding cost dominates. Also since $L_c \gg L_a$, for simplicity, ignore prompt and answer: let $L_q=0, L_a=0$ and $L=L_c$ be the context length.

We decompose compute into (1) \emph{core attention} (the $QK^\top$ and $\text{Attn}\cdot V$ matmuls), which scales as $O(L^2H)$, and (2) \emph{linear layers} (FFN + QKV/output projections), which scale as $O(LH^2)$. Let $H$ be the embedding dimension and $T$ be the number of layers. For GPT-2\footnote{Here we use GPT-2 architecture for simplicity. For other architecture like Qwen, only the constant term changes and the analysis still holds.} \citep{radford2019language} Transformer block, the core attention cost (number of multiplications) is $2TL^2H$ and the linear layers cost is $12TLH^2$.

For SuperThoughts, let $S$ be the number of superposition steps and let $L'$ be the CoT length, which includes both (1) steps when we superpose tokens and (2) steps when we use discrete tokens during adaptive inference. Note that without adaptive inference, $L'=S=L/2$ because we superpose at each CoT step. we break down the inference cost for each module:
\begin{itemize}
    \item \textbf{(Main module)} Core attention cost: $2TL'^2H$. Linear layers cost: $12TL'H^2$.
    \item \textbf{(MLP module)} Core attention cost: $2L'^2H$. Linear layers cost: $12L'H^2$. Additional projection cost: $3L'H^2$.
    \item \textbf{(Projection Compressor)} Projection cost: $2SH^2$.
    \item \textbf{(Transformer Compressor)} Transformer cost: $8SH+24SH^2$.
\end{itemize}
Note that for the Projection Compressor, the cost is $2SH^2$ not $2L'H^2$ because if we use adaptive inference and when we reject the MTP token, we do not compress tokens at the next CoT step. Ignoring the additional minor projection cost, a simple mental model could be that SuperThoughts has $\frac{L'}{L}$ CoT steps as the standard discrete model but each step costs $(T+1)/(T)$ compute. Note that since the core attention cost is quadratic in $L$ and $L'$, this can underestimate the compute saved depending on how large is $L$.

For SuperThoughts with Projection Compressor, ignoring these small projection-style overheads, a simple mental model is that SuperThoughts runs $L'$ decoding steps instead of $L$, but each step executes a $T$-layer backbone plus an extra 1-layer MTP, i.e., a per-step multiplier of $(T+1)/T$. This multiplier applies to both the dense-matmul term ($\propto LH^2$) and the core attention-mixing term ($\propto L^2H$). Approximating with the linear term gives
\[
\frac{C_{\text{ST}}}{C_{\text{base}}} \approx \frac{L'}{L}\cdot\frac{T+1}{T},
\]
which typically overestimates this ratio (and thus underestimates compute savings), since the core attention term scales as $\frac{T+1}{T}\left(\frac{L'}{L}\right)^2$.

\section{Algorithms for Dataset Preparation}\label{alg:dataset-prep}
\begin{algorithm}[ht]
\caption{SuperThoughts Dataset Preparation}\label{alg:soft-mtp}
\KwInput{Question $q$, answer $a$, chain token ids $\mathbf{c} = (c_1, \dots, c_N)$, token probabilities $\mathbf{p} = (p_1, \dots, p_N)$, window size $k$, special treatment mode $\mathcal{M} \in \{\texttt{prob}, \texttt{none}\}$}
\KwOutput{NTP tensors $(\mathbf{x}, \mathbf{G}, \mathbf{V}, \mathbf{y})$ and MTP tensors $(\mathbf{x}_{\text{mtp}}, \mathbf{G}_{\text{mtp}}, \mathbf{V}_{\text{mtp}}, \mathbf{y}_{\text{mtp}})$}
\BlankLine

\tcp{Step 1: Adaptive token selection (mode-specific, see Algorithms~\ref{alg:select-prob-frac}--\ref{alg:select-none})}
$\mathbf{s} \leftarrow \FAdaptiveSelect_{\mathcal{M}}(\mathbf{c}, \mathbf{p})$\;
\BlankLine

\tcp{Step 2: Window-aligned padding}
$\tilde{\mathbf{c}},\, \mathbf{m} \leftarrow \FWindowAlign(\mathbf{c}, \mathbf{s}, k, \mathcal{M})$\;
\BlankLine

\tcp{Step 3: Build NTP sequence}
$\mathbf{x},\, \mathbf{G},\, \mathbf{V},\, \mathbf{y} \leftarrow \FBuildSequence(q, a, \tilde{\mathbf{c}}, \mathbf{m}, k, \texttt{false})$\;
\BlankLine

\tcp{Step 4: Build MTP sequence (shift chain right by $k-1$)}
$\tilde{\mathbf{c}}_{\text{mtp}} \leftarrow (\underbrace{\textsc{cot\_pad}, \dots, \textsc{cot\_pad}}_{k-1},\, \tilde{c}_1, \dots, \tilde{c}_{|\tilde{\mathbf{c}}|})$\;
$\mathbf{m}_{\text{mtp}} \leftarrow (\underbrace{\texttt{true}, \dots, \texttt{true}}_{k-1},\, m_1, \dots, m_{|\mathbf{m}|})$\;
$\mathbf{x}_{\text{mtp}},\, \mathbf{G}_{\text{mtp}},\, \mathbf{V}_{\text{mtp}},\, \mathbf{y}_{\text{mtp}} \leftarrow \FBuildSequence(q, a, \tilde{\mathbf{c}}_{\text{mtp}}, \mathbf{m}_{\text{mtp}}, k, \texttt{true})$\;
\BlankLine
\Return $(\mathbf{x}, \mathbf{G}, \mathbf{V}, \mathbf{y}),\; (\mathbf{x}_{\text{mtp}}, \mathbf{G}_{\text{mtp}}, \mathbf{V}_{\text{mtp}}, \mathbf{y}_{\text{mtp}})$\;
\end{algorithm}

\begin{algorithm}[ht]
\caption{$\textsc{AdaptiveSelect}_{\texttt{prob}}$: Fraction-Based Probability Selection}\label{alg:select-prob-frac}
\KwInput{Token ids $\mathbf{c} = (c_1, \dots, c_N)$, token probabilities $\mathbf{p} = (p_1, \dots, p_N)$}
\KwOutput{Boolean mask $\mathbf{s} = (s_1, \dots, s_N)$}
\BlankLine
Sample $\alpha \sim \mathcal{U}[\alpha_{\min},\, \alpha_{\max}]$ \tcp*{Here $0\leq\alpha_{\text{min}} < \alpha_{\text{max}}\leq 1$ is a fraction}
$m \leftarrow \lfloor \alpha \cdot N \rfloor$\;
$\mathcal{I} \leftarrow$ indices of the $m$ smallest values in $\mathbf{p}$\;
$s_i \leftarrow \mathbb{1}[i \in \mathcal{I}] \quad \forall\, i \in \{1, \dots, N\}$\;
\Return $\mathbf{s}$\;
\end{algorithm}

\begin{algorithm}[ht]
\caption{$\textsc{AdaptiveSelect}_{\texttt{none}}$: No Selection (Baseline)}\label{alg:select-none}
$s_i \coloneqq \texttt{false} \quad \forall\, i$\;
\end{algorithm}

\begin{algorithm}[ht]
\caption{\textsc{WindowAlignedPad}: Window-Aligned Padding}\label{alg:window-align}
\KwInput{Token ids $\mathbf{c}$, special mask $\mathbf{s}$, window size $k$, mode $\mathcal{M}$}
\KwOutput{Padded ids $\tilde{\mathbf{c}}$, padding mask $\mathbf{m}$ ($m_i = \texttt{true} \Rightarrow$ inserted pad)}
\BlankLine

$\texttt{pad\_right} \leftarrow (\mathcal{M} \neq \texttt{prob-frac})$\tcp*{isolate specials in own window?}
$\tilde{\mathbf{c}} \leftarrow [\,],\quad \mathbf{m} \leftarrow [\,],\quad j \leftarrow 0$\tcp*{$j$: position within current window}

\BlankLine
\For{$i \leftarrow 1$ \KwTo $|\mathbf{c}|$}{
    \uIf{$s_i$}{
        \If(\tcp*[f]{pad to finish current window}){$j \neq 0$}{
            $\Append{[\textsc{cot\_pad}]^{k-j}}{[\texttt{true}]^{k-j}}$\;
        }
        $\Append{c_i}{\texttt{false}}$\;
        $j \leftarrow 1$\;
        \If(\tcp*[f]{fill rest of window}){$\texttt{pad\_right}$}{
            $\Append{[\textsc{cot\_pad}]^{k-1}}{[\texttt{true}]^{k-1}}$\;
            $j \leftarrow 0$\;
        }
    }
    \Else{
        $\Append{c_i}{\texttt{false}}$\;
        $j \leftarrow (j + 1) \bmod k$\;
    }
}
\Return $\tilde{\mathbf{c}},\, \mathbf{m}$\;
\end{algorithm}

\begin{algorithm}[ht]
\caption{\textsc{BuildSequence}: Assemble Input and Target Tensors}\label{alg:build-seq}
\KwInput{Question $q$, answer $a$, padded chain $\tilde{\mathbf{c}}$, padding mask $\mathbf{m}$, window size $k$, flag $\texttt{is\_mtp}$}
\KwOutput{Input ids $\mathbf{x}$, grouped CoT inputs $\mathbf{G} \in \mathbb{Z}^{L \times k}$, valid masks $\mathbf{V} \in \{0,1\}^{L \times k}$, targets $\mathbf{y}$}
\BlankLine

\tcp{Tokenize with chat template}
$\mathbf{t} \leftarrow \textsc{ChatTemplate}(q, a)$\tcp*{contains \texttt{<think>} $\dots$ \texttt{</think>}}
$b \leftarrow \text{index of } \texttt{<think>} \text{ in } \mathbf{t}$\;
$e \leftarrow \text{index of } \texttt{</think>} \text{ in } \mathbf{t}$\;
\BlankLine

\tcp{Group padded chain into $k$-windows}
$G_l \leftarrow (\tilde{c}_{(l-1)k+1}, \dots, \tilde{c}_{lk})$ for $l = 1, \dots, L_{\text{cot}}$\;
$V_l \leftarrow (\lnot\, m_{(l-1)k+1}, \dots, \lnot\, m_{lk})$ for $l = 1, \dots, L_{\text{cot}}$\;
\BlankLine

\tcp{Construct input ids: replace reasoning span with placeholders}
$\mathbf{x} \leftarrow [\,\mathbf{t}_{1:b},\; \underbrace{\textsc{pad}, \dots, \textsc{pad}}_{L_{\text{cot}}},\; \mathbf{t}_{e:|t|}\,]$\;
$r_{\text{start}} \leftarrow b + 1,\quad r_{\text{end}} \leftarrow b + L_{\text{cot}}$\;
$\mathbf{G}[r_{\text{start}} : r_{\text{end}}] \leftarrow (G_1, \dots, G_{L_{\text{cot}}})$\tcp*{elsewhere filled with \textsc{pad}}
$\mathbf{V}[r_{\text{start}} : r_{\text{end}}] \leftarrow (V_1, \dots, V_{L_{\text{cot}}})$\tcp*{elsewhere \texttt{false}}
$\texttt{cot\_mask}_i \leftarrow \bigvee_{j=1}^{k} V_{i,j}$\tcp*{true if position $i$ has any valid CoT token}
\BlankLine

\tcp{Construct targets}
\For{$i \leftarrow 1$ \KwTo $L$}{
    \uIf{$\texttt{cot\_mask}_i$}{
        $\mathbf{y}_i \leftarrow$ first $G_{l(i)+1}[j]$ such that $G_{l(i)+1}[j] \neq \textsc{cot\_pad}$\tcp*{next window's first valid token}
    }
    \uElseIf{$\texttt{is\_mtp}$ \textbf{or} $i \leq r_{\text{end}}$}{
        $\mathbf{y}_i \leftarrow \textsc{ignore}$\tcp*{mask out question region; MTP masks answer too}
    }
    \Else{
        $\mathbf{y}_i \leftarrow \mathbf{x}_{i+1}$\tcp*{standard next-token prediction for answer}
    }
}
\If{$\lnot\, \texttt{is\_mtp}$}{
    $\mathbf{y}_b \leftarrow G_1[1]$\tcp*{at \texttt{<think>}, predict first CoT token}
}
\BlankLine
\Return $\mathbf{x},\, \mathbf{G},\, \mathbf{V},\, \mathbf{y}$\;
\end{algorithm}

\end{document}